\documentclass[10pt, a4paper]{article}

\usepackage[]{lrec-coling2024} % this is the new style

\usepackage{times}
\usepackage{latexsym}
\usepackage{enumitem}
\usepackage{textcomp}
\usepackage[T1]{fontenc}
\usepackage[utf8]{inputenc}

\usepackage{microtype}

\usepackage{graphicx}
\usepackage{hyperref}
\usepackage{csquotes}
\usepackage{xcolor}
\usepackage{adjustbox}
\usepackage{makecell}
\usepackage{multirow}
\usepackage{amsmath}
\usepackage{amssymb}
\usepackage{pifont} %for xmarks
\title{Modeling Low-Resource Health Coaching Dialogues via Neuro-Symbolic Goal Summarization and Text-Units-Text Generation\\ \vspace*{.5\baselineskip}} 

\name{Yue Zhou, Barbara Di Eugenio, Brian Ziebart, Lisa Sharp, \\ {\bf \large Bing Liu and Nikolaos Agadakos } } 

\address{University of Illinois Chicago, Illinois, United States \\
         \{yzhou232, bdieugen, bziebart, sharpl, liub, nagada2\}@uic.edu\\
         }

\abstract{
Health coaching helps patients achieve personalized and lifestyle-related goals, effectively managing chronic conditions and alleviating mental health issues. It is particularly beneficial, however cost-prohibitive, for low-socioeconomic status populations due to its highly personalized and labor-intensive nature. In this paper, we propose a neuro-symbolic goal summarizer to support health coaches in keeping track of the goals and a text-units-text dialogue generation model that converses with patients and helps them create and accomplish specific goals for physical activities. Our models outperform previous state-of-the-art while eliminating the need for predefined schema and corresponding annotation. We also propose a new health coaching dataset extending previous work and a metric to measure the unconventionality of the patient's response based on data difficulty, facilitating potential coach alerts during deployment. % are also proposed.
 \\ \newline  }

\begin{document}

\maketitleabstract

\section{Introduction}

Health coaching is a patient-centered clinical practice that aims to help patients achieve personalized and lifestyle-related goals to enhance their health behaviors. It has demonstrated efficacy in managing chronic conditions like diabetes and cardiovascular disease, as well as alleviating mental health issues such as anxiety and depression~\citep{hc4butterworth2006effect,hc1ghorob2013supplement,hc3kivela2014effects,hc2thom2016qualitative}. Health coaching is particularly advantageous for low-socioeconomic status (SES) populations, who endure a disproportionate burden of physical and mental health issues~\citep{ses2thackeray2004disparities,ses1kangovi2014challenges}. Nonetheless, it is invariably cost-prohibitive for these populations due to its highly personalized and labor-intensive nature; additionally, it also requires considerable time commitment since it normally spans several sessions across weeks or even months. 

Recently, efforts have been undertaken to enhance the effectiveness of health coaching through natural language processing techniques. Our group~\citep{gupta1-etal-2020-human,gupta22020goal,gupta32021summarizing} collected real-world datasets of health coaching conversations and proposed a goal summarization approach based on slot-filling and dialogue state tracking. We further improved goal summarization and introduced an efficient health coach dialogue system that supports empathetic response generation~\citep{zhou-hc-22}. An excerpt of a health coaching conversation is illustrated in Table~\ref{tab:teaser}.

% \begin{figure}[!t]
%     \centering 
%     %\def\svgscale{0.65}
%     %\def\svgwidth{\columnwidth}
%     \includegraphics[width=\columnwidth]{ts.pdf}
%     \caption{An excerpt from a health coaching conversation from our dataset. Each week the coach and patient collaboratively establish a feasible goal. Then, the coach monitors the patient's progress, maintains the patient's engagement, and addresses the patient's concerns.}
%     \label{fig:teaser}
% \end{figure}
\begin{table*}[h]
\begin{adjustbox}{width=\textwidth}
\begin{tabular}{l}
       \\
\hline\hline
\textbf{Excerpt of Patient-Coach Conversation~\#~17 Week~\#~7} \\ \hline
\makecell[cl]{
Coach: Hi [Name]! I hope you had a great weekend :) You are now in the second to last week of the study! \\
Coach: Your goal last week was to reach 7000 steps everyday - you got close! \\
Coach: Did you want to keep your walking goal the \textbf{same this week or lower it a bit?} \\
Patient: You can lower it to 5000 \\
Coach: Okay, sounds good - and what days are you thinking to do 5,000 steps? \\
Patient: 7 days \\
Coach: awesome - and what's your confidence on a scale from 1-10 that you can reach this goal?}     \\    
\hspace{3cm}... \hspace{3cm} ... \hspace{3cm} ...\\
\makecell[cl]{
Coach: Hi [Name] - you're doing a really great job with your goal this week. \\
Coach: You got over 11,000 steps on Monday and yesterday - that's over double your goal! Amazing job :) \\
Patient: Thanks and you too \\
Coach: Hi [Name] - Great job reaching your goal last week. You got over 5,000 steps all 7 days! You should be really proud of yourself :)} \\\hline
\textbf{Excerpt of Patient-Coach Conversation~\#~17 Week~\#~8} \\ \hline

\makecell[cl]{
Coach: This is your last week in the study - what would you like your final walking goal to be? \\
Patient: Stay the \textbf{same as last week}.} \\
\hspace{3cm}... \hspace{3cm} ... \hspace{3cm} ...

\\\hline
\end{tabular}
\end{adjustbox}
\caption{An excerpt from a health coaching conversation from our dataset. Each week the coach and patient collaboratively establish a feasible goal. Then, the coach monitors the patient's progress, maintains the patient's engagement, and addresses the patient's concerns. The coach may discuss the goal by referring to the goal settled before.}
\label{tab:teaser}
\end{table*}

However, these works are based on simplified dialogue states (a set of goal attribute-value pairs) and require a designed schema with corresponding human annotations, which are not only labor-intensive but also lose the global contextual information that is established over several sessions (in our case, from 4 to 8 weeks). In addition, the considerable length of each sample, together with the limited data size, poses significant challenges in healthcare dialogue modeling. To address these issues, we propose a neuro-symbolic goal summarization approach that eliminates the need for a pre-designed schema or corresponding annotations yet preserves interpretability. The model is optimized to (i) summarize the goal as effectively as possible, given the current week's dialogue, and (ii) generate an executable instruction on how to modify the summarized goal by referencing the goal from the previous week(s). We also introduce a text-units-text dialogue generation approach that considers as input a sequence of discrete units symbolizing the dialogue history. Moreover, we propose a more generalized approach to detect patient's unconventional responses by extending Point-wise $\mathcal{V}$-usable Information (PVI)~\cite{PVI} for dialogue generation, without using any external datasets, contrasting with our previous work~\citep{zhou-hc-22}. Finally, we introduce a novel health coaching dataset consisting of 1880 dialogue turns from 22 patient-coach conversations, each spanning up to 8 weeks. The dataset is enriched with Fitbit data, which tracks the progress of patients' goals. We aim to augment the existing health coaching datasets and provide a more robust testing benchmark for health coaching modeling, particularly in potential domain-shift scenarios.

We evaluate our model by both automatic metrics and expert-based human evaluation. Experimental results show that our neuro-symbolic goal summarizer outperforms the current state-of-the-art by up to $\sim$30\% in semantic frame accuracy. In addition, our text-unit-text dialogue generation achieved the best performance compared to previous work in all metrics for all datasets. Health coaches prefer our generated responses over the previous state-of-the-art 33.9\% of the time, while the vice versa happens 19.6\% (the rest are ties). 

Our contributions are (1) a data-efficient neuro-symbolic goal summarization model and text-units-text dialogue generation model for health coaching, which outperform previous state-of-the-art while eliminating the need for predefined schema and corresponding annotation; (2)  a metric measuring the unconventionality of the patient's response in terms of data difficulty, facilitating potential coach alert during deployment and data characterization during training; and (3) a novel health coaching dataset.

\section{Related Work}

\paragraph{Conversational Agents in Healthcare.} Conversational agents have been employed in healthcare to enhance the efficiency and scalability of interactions between healthcare professionals and patients. For instance, chatbots have been utilized in various healthcare contexts, such as chronic disease monitoring~\citep{sms3chaix2020vik}, cognitive behavior therapy~\citep{sms2fitzpatrick2017delivering}, and physical activity promotion~\citep{sms4mohan2020designing, sms5kocielnik2018reflection}. Nevertheless, these systems often exhibit limitations in their natural language understanding and generation capabilities. More advanced approaches have been proposed for mental health counseling~\citep{consolalthoff-etal-2016-large, consol2shen-etal-2020-counseling} and health coaching. Our group \citep{gupta1-etal-2020-human, gupta22020goal, gupta32021summarizing} collected two real-world health coaching conversation datasets, focusing on the NLU components that summarize weekly goals to support health coaches. Building upon these datasets, we developed a data-efficient health coaching dialogue system with a simplified NLU and NLG framework and mechanism-conditioned empathetic response generation~\citep{zhou-hc-22}. In recent years, there has been a shift in domain-specific dialogue systems from modularized and NLU-NLG component-based designs~\citep{jokinen09, tod1williams2016dialog, tod2budzianowski-etal-2018-multiwoz, tod3mrkvsic2016neural, tod4wen-etal-2015-semantically} towards end-to-end architectures to reduce human effort and error propagation between modules~\citep{tode2e3hosseini2020simple, tode2e4peng2020soloist}.

\paragraph{Neuro-symbolic Approaches} have recently gained significant attention due to their ability to facilitate end-to-end while leveraging symbols for interpretability and data-efficient training. \citet{Mao2019} proposed a neuro-symbolic concept learner that combines the strengths of both neural networks and symbolic logic, demonstrating improved performance in visual question-answering tasks. \citet{Lamb2019} introduced the Neuro-Symbolic Transformer, which leverages symbolic reasoning within a transformer-based architecture in text classification tasks. \citet{deraedt2019neurosymbolic} proposed a neuro-symbolic approach that demonstrates the potential of neuro-symbolic systems in handling complex reasoning tasks. Another work in this area is the exploration of end-to-end differentiable natural logic modeling \cite{feng-etal-2020-exploring}. \citet{dong-lapata-2018-coarse} proposed a coarse-to-fine decoding method for neural semantic parsing, utilizing a structure-aware neural architecture. \citet{nsconv} explores a neuro-symbolic approach for enhancing conversational dialogue systems with commonsense reasoning capabilities. \citet{garcez2018neurosymbolic} provides a comprehensive survey and taxonomy of approaches that combine neural and symbolic methods for learning and reasoning in AI systems.

\section{Health Coaching Datasets}

\paragraph{Background}
A health coaching process often starts with a goal-setting stage where the coach discusses creating a S.M.A.R.T. goal with the patient, namely a goal that is specific, measurable, achievable, relevant, and time-bound \cite{doran1981there}.
%in a motivational interviewing manner. 
Once the goal is settled, the coach will monitor the patient's progress and maintain patient engagement. Our research group~\citep{gupta1-etal-2020-human} collected two datasets (dataset~1 and~2) of health coaching dialogues between patients and coaches via text messages. 
Previously, we defined ten slots for the goal's attributes (types of activity, amount, time, days, location, duration, and the confidence score for the activity). We also used a stage-phase schema for additional turn-level annotation and added dialogue act annotations in our later work~\citep{gupta32021summarizing}. 

\paragraph{New Human Data Collection}
Emulating their settings, we recruited four health coaches trained in SMART goal setting and a cohort of 22 patients. The study was conducted over eight weeks. The communication between the coach and the patient was facilitated through text messaging applications, and Fitbits were provided to monitor patients' activity progress. However, six patients withdrew in the early weeks of the study. Consequently, the dataset comprised 1880 dialogue turns over 102 weeks. The datetime of messages, interlocutors, dialogue content, Fitbit action data, and patient and health coach IDs are available for each dialogue. The data were thoroughly anonymized. In contrast to the first two rounds of data, we retained the emoji tokens in the dataset for future analysis. Data statistics are shown in~\ref{tab:ds1}. A sample of the health coaching conversation between the patient and health coach is presented in Table~\ref{tab:teaser}. The full dataset is available at \href{https://github.com/uic-nlp-lab/virtualcoachdata/}{https://github.com/uic-nlp-lab/virtualcoachdata/}.
\section{Methods}

\begin{table}
    \centering
    \begin{adjustbox}{width=0.85\columnwidth}
    \begin{tabular}{lllll}
    \hline\hline
                                                       & \textbf{\#P/C}  & \textbf{\#W}   & \textbf{\#T}     & \textbf{\#T/(W,P)} \\ \hline
\begin{tabular}[c]{@{}l@{}}
Dataset~1\end{tabular} & 27/1 & 4   & 2,853 & 26.6  \\ 
\begin{tabular}[c]{@{}l@{}}Dataset~2\end{tabular} & 28/3 & 5-8 & 4,134 & 18.8  \\ 
\begin{tabular}[c]{@{}l@{}}Ours\end{tabular} & 22/4 & 2-8 & 1,880 & 18.4 \\ \hline
\end{tabular}
\end{adjustbox}
    \caption{Dataset Statistics. \#P/C, \#W, \#T, \#T/(W,P) refer to the number of patients/health coaches involved, study weeks, total dialogue turns, average turns per patient per week. }
    \label{tab:ds1}
\end{table}

% Table~\ref{tab:ds2} presents a patient and health coach dialogue from our collected data.

% \begin{table*}[!h]
% \begin{adjustbox}{width=\textwidth}
% \begin{tabular}{lll}
% \hline
% \hline
% dt & \textbf{Dialogue Context}  & \textbf{Fitbit}                        \\ \hline
% dt1 & \makecell[cl]{Coach: and one last thing, how would you rate the goal? }                            & 5000          \\

% dt2 & \makecell[cl]{Patient: and one last thing, how would you rate the goal? }                            & 6000          \\\hline

% \end{tabular}
% \end{adjustbox}
% \caption{example real}
% \label{tab:ds2}
% \end{table*}
% A sample of the health coaching conversation between the patient and health coach is presented in Figure~\ref{fig:teaser}.

This section presents the two integral components of our health coaching dialogue system: the Neuro-Symbolic Goal Summarizer and the Neuro-Symbolic Dialogue Generator.

%, dataset description, and evaluation metrics.
\subsection{Neuro-Symbolic Goal Summarizer}\label{nssum}

The task of goal summarization involves processing the dialogue history between the health coach and the patient to generate the negotiated goal in natural language. The dialogues in health coaching often span multiple weeks. When the coach discusses the current week's goal, they may refer back to the attributes of the goal implemented in the previous week. However, incorporating all text from previous weeks can result in excessively long inputs, which, given the limited dialogue samples, can be challenging to train on. Our previous research \citep{zhou-hc-22, gupta32021summarizing} approached this task by formulating it as dialogue state tracking, where the state is defined as a set of goal attribute-value pairs. However, these methods require a pre-defined schema with corresponding annotations and a separate slot-filling module, which do not optimally utilize the entire context, resulting in compromised information.

To address these challenges, we aim to provide a data-efficient and end-to-end yet interpretable solution and propose a neuro-symbolic approach for goal summarization. The approach considers the dialogue of the current week while referencing the previous week's goal. In addition, it does not require a schema or corresponding annotations. We aim to optimize the model to summarize the goal, focusing on two key aspects: (i) summarizing the goal as effectively and comprehensively as possible, given the \emph{current} dialogue, and (ii) generating a feasible instruction on how to modify the summarized goal from (i) based on the previous goal. As an example depicted in Figure~\ref{fig:nsg}, the coach elaborates on the goal for week $w_t$ by referencing the goal established in the preceding week $w_{t-1}$ (\emph{``same days?''}). Our model learns to generate an executable instruction (\texttt{Copy \{Days\}}) and extract the partial goal (\emph{``Walk 2,500 steps''}) from the current week's dialogue, and execute the instruction on the referenced previous week's goal to generate the complete goal summarization: \emph{``Walk 2,500 steps from Monday to Friday.''} 

To achieve that, we formulate the problem within the context of Reinforcement Learning (RL), specifically utilizing the Proximal Policy Optimization (PPO)~\citep{PPO} framework. We maximize the following training objective:

\begin{figure*}[!htb]
    \centering 
    \includegraphics[width=0.85\textwidth]{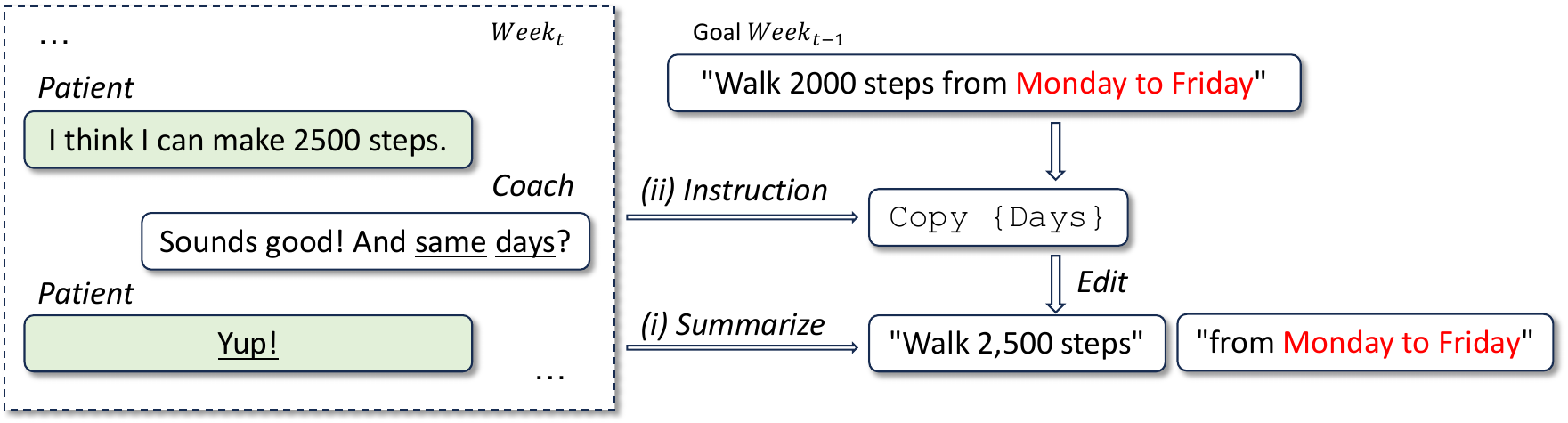}
    \caption{A simplified demonstration of Neuro-Symbolic Goal Summarization. The health coach discusses the goal for week $w_t$ by referring to the goal set in the previous week $w_{t-1}$ (\emph{"same days?"}). Our model is trained to generate an executable instruction (\texttt{Copy \{Days\}}) and to extract the partial goal (\emph{"Walk 2,500 steps"}) from the dialogue of the \emph{current} week. The model then edits the partial goal by applying the instruction to the reference previous goal, resulting in the comprehensive goal summary: \emph{"Walk 2,500 steps from Monday to Friday."}}
    \label{fig:nsg}
\end{figure*}

\begin{multline}
\text{objective}(\phi) = \mathbb{E}(x, y)\sim  D_{\pi_{\phi}^{RL}} [r(y,y^*|x)  \\- \lambda \text{ KL}( \pi_{\phi}(y|x) || \pi_{base}(y|x) )]
\end{multline}

\noindent where $\pi_{\phi}$ is the learned RL policy (e.g., the language model to be fine-tuned) and $\pi_{base}$ is the untouched pre-trained language model. $r$ is the reward function (we use ROUGE score~\citep{rouge}), and the KL term is the Kullback–Leibler Divergence as regularization. We aim to optimize the likelihood of the generated sequences, comprising partial goals and instructions, that \texttt{Instruction\{Partial Goal, Reference\}} subsequently results in ground truth summarization. The full set of instructions (e.g., \texttt{Add \{Num\}}, \texttt{Copy \{Times\}}) will be available in the Appendix. 

\subsection{Text-Units-Text Dialogue Generator}

\subsubsection{Generation with Unit Symbols}

A prevalent issue in healthcare dialogue datasets is the extensive length of each sample, associated with the limited number of samples. Training a sequence-to-sequence generation model using a long context window presents a significant challenge. However, an interesting observation is that healthcare dialogues typically follow a similar pattern. For instance, health coaching invariably begins with the coach discussing a realistic goal with the patient, confirming each aspect of the goal, such as activity, amount, day, and times. Once the goal is established, the coach monitors the patient's progress and sustains patient engagement. 

Inspired by this observation and the recent success of speech-to-unit modules that predict the discrete representations of the speech~\citep{unit1,unit2}, we explore the possibility of utilizing a short context window while preserving the dialogue history information through symbolic abstraction in text-to-text dialogue generation. Specifically, our response generation incorporates the following two models:

\paragraph{Dialogue-History-to-Unit Encoder} encodes the long dialogue history into discrete unit symbols using out-of-box pre-trained language models and unsupervised approaches. We first encode each turn of the conversation in the training dialogues with SBERT, then run k-means clustering to obtain K clusters; we subsequently designate the cluster indices as units. Despite the unique details of each dialogue, we anticipate a similar sequence of units to emerge as each dialogue progresses. Furthermore, the total number of units in a dialogue can indicate the dialogue's progression length. 

\paragraph{Units-to-Text Generation} We propose a sequence-to-sequence model that generates responses for a virtual coach. The model takes as inputs a sequence of discrete units $N_1, ..., N_{t-1}$ symbolizing the dialogue history, the most recent dialogue turns from the coach $C_{t-1}$ and the patient $U_{t-1}$, and a partial goal  $G_{t-1}$ summarized as discussed in Section~(\ref{nssum}) based on the current dialogue. All inputs are concatenated as a single sequence. Then the generated virtual coach response $R_t$ is defined as:

\begin{equation}
R_t = \textrm{Seq2Seq}([N_1, ..., N_{t-1},G_{t-1}, R_{t-1},U_{t-1}])
\label{the 3}
\end{equation}

\subsubsection{Measuring Data Difficulty of User Responses}

Identifying unconventional user responses is vital in domain-specific dialogue systems to circumvent system failures due to unsupported user inputs. It can be particularly critical in healthcare settings, where patients often express concerns and emotions that exceed the system's capabilities or do not respond directly to the system's prompts. The unconventional inputs from patients, coupled with the limited availability of healthcare training data, can pose significant challenges to the model's training process. In this work, we show the potential of leveraging data difficulty to identify unusual patient inputs. 

Data difficulty metrics categorize data into easy-to-learn and difficult-to-learn examples based on training dynamics~\citep{AUM, carto} or mutual information~\cite{pvi-thoery,PVI}. Previous research primarily focuses on classification tasks, where a difficult-to-learn example typically signifies that the example is mislabeled. \citet{pvi-thoery} propose a framework called \textbf{predictive $\mathcal{V}$-information} to measure how much information can be extracted or learned from input $X$ about its label $Y$ when constrained to functions or a model family $\mathcal{V}$. The predictive $\mathcal{V}$-information, $I_{\mathcal{V}}(X\!\rightarrow\!Y)$ is defined as the difference between \textit{predictive $\mathcal{V}$-entropy} $H_{\mathcal{V}}(Y)$ and \textit{conditional $\mathcal{V}$-entropy} $H_{\mathcal{V}}(Y|X)$:
\begin{equation}
% H(x->y) = placeholder 7
I_{\mathcal{V}}(X\!\rightarrow\!Y)=H_{\mathcal{V}}(Y)-H_{\mathcal{V}}(Y|X)
\label{eq:pvi3}
\end{equation}
The greater the $I_{\mathcal{V}}(X\!\rightarrow\!Y)$, the easier the dataset is for $\mathcal{V}$. 
Extending from~\citet{pvi-thoery}, ~\citet{PVI} propose \textbf{Point-wise $\mathcal{V}$-usable Information (PVI)} to measure the information in the \emph{individual} instance usable by a model family $\mathcal{V}$ with respect to the data distribution. The PVI of an instance $(x,y)$ is defined as:
\begin{equation}
% PVI = placeholder 8
%\operatorname{PVI}(x \rightarrow y)=-\log _2 g_{y}(\varnothing)+\log _2 g^{\prime}_{y}(x)
\operatorname{\mathrm{PVI}}(x\!\rightarrow\!y)=-\log _2 g^{\prime}_{y}(\varnothing)+\log _2 g_{y}(x)
\label{eq:pvi4}
\end{equation}
where $g^{\prime} \in \mathcal{V}$ s.t. $\mathbb{E}[-\log g^{\prime}_{Y}(\varnothing)]=H_{\mathcal{V}}(Y)$ and $g \in \mathcal{V}$ s.t. $\mathbb{E}\left[-\log g_{Y}(X)\right]=H_{\mathcal{V}}(Y|X)$. If $\mathcal{V}$ were the BERT function family, $g$ and $g^{\prime}$ would be the models after finetuning BERT with and without the input, respectively. An instance with a large negative PVI value is considered as ``difficult,'' showing that the model can better predict the majority class when ignoring $X$, which often indicates the instance is mislabeled. 

Inspired by their work, we aim to explore the difficulty in predicting user responses given the most recent dialogue context, which will provide insights into the degree of unconventionality or surprise in the user's response and its divergence from the dataset distribution. This approach could additionally serve as an easy-to-implement and general out-of-domain detection, eliminating the need to construct a separate domain-specific detection classifier (e.g., empathy detector~\citep{zhou-hc-22}). To achieve this, we propose an \emph{extension to the PVI} in sequence-to-sequence \emph{generation} as:

\begin{multline}
\operatorname{\mathrm{PVI}}(x\!\rightarrow\!y) = -\underbrace{\sum_{t=1}^{n}log(p(y_t|y_{<t},\varnothing))}_{\text{by } g{}'} \\ + \underbrace{\sum_{t=1}^{n}log(p(y_t|y_{<t},x))}_{\text{by } g} 
\end{multline}

\noindent where $x$ and $y$ are the sequence of the dialogue context and the user response, and $g$ and $g^{\prime}$ are the generative models after fine-tuning with and without the dialogue context, respectively. An instance (user response) with a large negative PVI value is considered ``difficult,'' showing that the model can better predict the user response when ignoring the context, indicating the unconventionally in the user response. 

The difficulty metric we propose serves two functions: Firstly, it assists in identifying unconventional patient responses during deployment. If such a response were detected, the system could alert the human coach, indicating potential patient concerns that require attention or suggesting that the patient's utterance or question may not be within the system's capabilities. Secondly, the difficulty metric acts as a data filter during training. It categorizes our limited data into subsets of easy-to-learn and hard-to-learn instances, facilitating possible curriculum learning or denoising approaches for our system.

\section{Experiments}

\subsection{Experimental Settings}

To ensure comparability, we use dataset~1 and dataset~2 from ~\citet{gupta32021summarizing} for training/development and testing, following our previous work~\citep{zhou-hc-22}. We also employ identical model architecture backbones as before, specifically utilizing T5~\cite{t5-raffel2019exploring} for the goal summarization task and GPT-2~\cite{gpt2-radford2019language} for the response generation task. We additionally evaluate model response generation performance on our newly collected dataset~3 for generalizability in possible domain-shifting. To mitigate inefficient sampling, we manually annotate 40 positive examples. We also use two T5-base models for our PVI-generation metrics. We choose $k$ = 15 for deriving the discrete units. Further details and model parameters will be available in the Appendix.

\paragraph{Evaluation Metrics} We use semantic frame correctness for evaluating the goal summarization, which is identical to goal correctness@$k$ with $k$ = 10, used by our previous work~\cite{gupta32021summarizing,zhou-hc-22}. For dialogue response generation, we use BLEU~\citep{bleu-papineni-etal-2002-bleu}, BertScore~\citep{bert-score}, and Perplexity (PPL). We measure fluency as perplexity (PPL) of the generated response using a pre-trained GPT2 model that has not been fine-tuned for this task, following previous work~\citep{ppl-ma-etal-2020-powertransformer,emp-rewriting10.1145/3442381.3450097}.

.

\subsection{Results}

We compare our Neuro-symbolic Summarization model with the best performance reported by~\citet {gupta32021summarizing} and ~\citet{zhou-hc-22} in Table~\ref{tab:res1}. D1 (F), D1 (B), and D2 refer to dataset~1 (Forward), dataset~1 (backward), and dataset~2. Forward and Backward refer to the two points where they labeled the goals for each week: one at the end of the goal-setting
stage (forward) and the other at the end of the goal-implementation stage (backward). Our approach improves semantic frame correctness by a significant margin ($\sim$ 20$\%$ - $\sim$30$\%$), compared with previous work that utilizes the traditional slot-filling and state-tracking framework.

\begin{table}
    \centering
    \begin{adjustbox}{width=0.85\columnwidth}
    \begin{tabular}{llll}
    \hline\hline
            \textbf{Model}  & \textbf{D1 (F)} & \textbf{D1 (B)} & \textbf{D2} \\ \hline
\citet{gupta32021summarizing}            & 15            & 13.1          & -         \\
\citet{zhou-hc-22}           & 21.7          & 31.6          & 13.6      \\
Ours & \textbf{52.3}          & \textbf{44.8}          & \textbf{46.7}   \\ \hline 
\end{tabular}
\end{adjustbox}
    \caption{Goal summarization performance by semantic frame correctness. D1 (F), D1 (B), and D2 refer to dataset~1 (Forward), dataset~1 (backward), and dataset~2. Our model outperforms previous work by a significant margin in all three datasets.}
    \label{tab:res1}
\end{table}

Table~\ref{tab:res2} shows the model performance on dialogue response generation. We employ two settings to train our response generation model: (1) Original. Using the original data, and (2) Low-PVI-Replace. substituting the patient's response, which has been assigned a large negative PVI value, with an alternative response from a similar context with a positive PVI in the training set. This context is identified by locating the semantically closest previous coach utterance using sentence-BERT embeddings. We substitute 5\% of the patient's utterances, prioritizing those with the most negative PVI values in the training set. We acknowledge that patient utterances with low negative PVI values are not necessarily "noise," and such substitutions could potentially disrupt the coherence of subsequent dialogues. However, we have found that this replacement strategy generally aids our model's convergence, particularly with limited training data. Our model (both with and without PVI fix) outperforms previous work on both dataset 2 and our newly collected dataset in all metrics. We have identified a domain shift in our collected dataset, characterized by patients exhibiting less verbal communication and increased usage of emojis. Coaches have adapted to maintain the continuity of the dialogue by becoming more talkative. The domain shift negatively impacted our model's performance.
\begin{table*}[h]
\centering
\begin{adjustbox}{width=0.85\textwidth}

\begin{tabular}{lllllll}
\hline\hline
& & \textbf{Data~2} & & & \textbf{Data~3} & \\
\hline
Model   & BLEU  ($\uparrow$)        & PPL ($\downarrow$)           & BertS F1 ($\uparrow$) & BLEU  ($\uparrow$)        & PPL ($\downarrow$)           & BertS F1 ($\uparrow$) \\ \hline
\citet{zhou-hc-22}     & 25.1         & 15.6         & 87.2  & 20.6         & 19.7         & 83.5            \\
OURS+Original    & 26.7 & 14.7 & 88 & 22.5 & 16.7 & \textbf{84.4} \\
OURS+Low-PVI-Replace    & \textbf{28.4} & \textbf{14.1} & \textbf{89.2} & \textbf{23.7} & \textbf{15.3} & 84.1\\
\hline
\end{tabular}

\end{adjustbox}
\caption{Evaluation on dialogue generation. BLEU: Average of BLEU-1,-2,-3,-4. BertS F1: Averaged BERT-score F1. Our model (both with and without PVI fix) outperforms previous work on both dataset 2 and our newly collected dataset 3 in all metrics. Nevertheless, the performance of all models diminishes on dataset 3 due to possible distribution shifts.}
\label{tab:res2}

\end{table*}

\paragraph{Human Evaluation}

\begin{figure}[!ht]
\begin{center}
\includegraphics[scale=0.5]{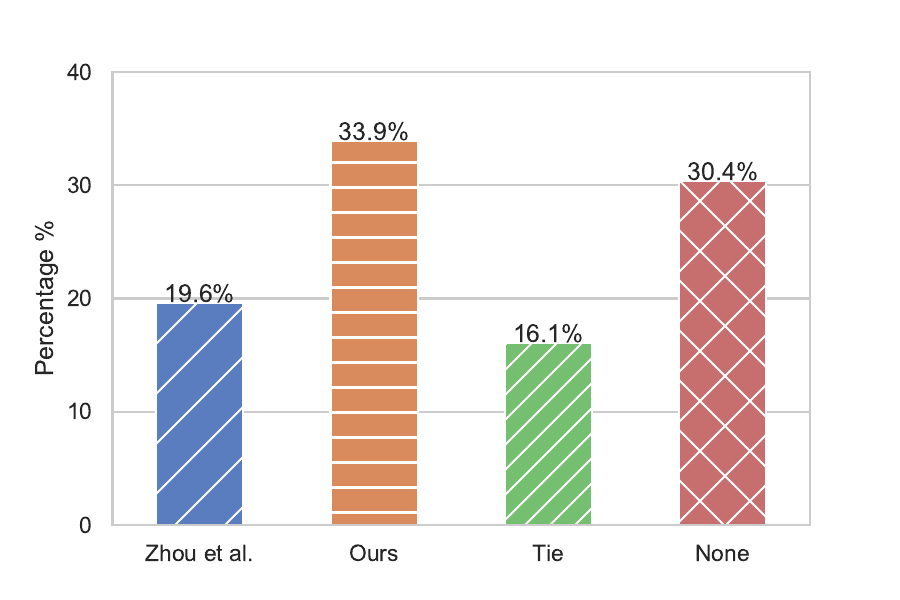} 
\caption{Human evaluation of generated response by health coaches. Our model's outputs have a $14.29\%$ preference over previous state-of-the-art~\cite{zhou-hc-22}.}
\label{hev}
\end{center}
\end{figure}

We conducted an expert-led assessment of the generated coach responses via A/B testing. Four health coaches were consulted to evaluate our model's outputs compared to those from previous state-of-the-art~\cite{zhou-hc-22}, given the same dialogue histories. The coaches were asked to select one of the following options: (a) Response A is most suitable in the given context; (b) Response B is most suitable in the given context; (c) Tie; both responses are equally good, or (d) Neither is appropriate. Among 56 collected samples, our model's outputs have a 14.29$\%$ preference over previous state-of-the-art~\cite{zhou-hc-22} (as the difference between the preferences for one or the other). However, approximately 30$\%$ of the generated responses were not deemed appropriate by our experts. This indicates the ongoing challenge of generating high-quality health coaching responses, particularly given the constraints of limited data and the sensitive nature of health-related interactions. See Figure~\ref{hev}. 

\subsection{Qualitative Analysis}

We present the output examples from our neuro-symbolic goal summarizer and dialogue generator, comparing them to previous work. We also demonstrate generation from GPT-3.5-turbo\footnote{https://platform.openai.com/docs/guides/chat} (two-shot in-context exemplars due to the input context limit), highlighting the limitation of utilizing LLMs in low-resource healthcare settings.

%the existing misalignment between our datasets and the health coaching scenario as perceived by the large Language Models.

Table~\ref{tab:qa1} compares goal summarization results from various models given the same dialogue history example. The dialogue example contains misleading information, such as \emph{``Friday''} and \emph{"2 miles a day"} (highlighted in red in the Table), while the part of the ground-truth information is in the previous goal (\emph{"7 days a week"}), as indicated by the bolded text  \emph{"same as last week"} that coach and patient agree to. Our previous methodologies, which rely on slot-filling and update dialogue states as the goal by either the last appearance of the slot-value~\cite{gupta32021summarizing} or a carryover classification~\cite{zhou-hc-22} and neglect the global context, fail to extract the correct goal information. In contrast, not only does our model summarize \emph{"3 miles"} from the dialogue but also generates the instruction of {\it copy(days)}, applying this instruction to the previous goal to complete the summarization with \emph{"7 days a week."} Moreover, our model delivers outputs in a natural language format, which is more user-friendly and comprehensible, particularly for health coaches, compared to the slot-values format. Lastly, the result from GPT3.5-turbo only captures partial information but asserts that the goal is a well-defined SMART goal.

\begin{table*}[h]
\begin{adjustbox}{width=\textwidth}
\begin{tabular}{l}
       \\
\hline\hline
\textbf{Dialogue  +  {[}Previous Goal: {walk 2 miles a day \textcolor{blue}{7 days a week}.]} } \\ \hline
\makecell[cl]{
Coach: Also, what would you like to set as your SMART goal this week? \\
Patient: Hey bro last \textcolor{red}{Friday} I met my goal but I didn't have the Fitbit on. \\
Patient: On \textcolor{red}{Tuesday} I had it on but it went dead. I met my goal on that day also. My goal for this week is \textcolor{red}{2 miles a day}. \\
Coach: No problem, making the goal is most important. Your body is your tracker.\\
Patient: Ok I'm excited and thanks for your help. \\
Coach: Okay so you want to \textbf{stick with your same goal} as last week? FYI \textcolor{red}{2 miles is 4000 steps}. \\
Patient: Let's make it \textcolor{blue}{3 miles}.\\ 
Coach: Base on your steps the first week I think 3 miles a day \textbf{is doable} if we monitor your step and have a plan... \\
Coach: \textbf{How many days do you want to go for? }\\
Patient: \textbf{Same as last week? }\\
Coach: \textbf{Sure}. And last what might get in the way of you accomplishing your goal? \\
Patient: Nothing. Patient: I'm ready. \\
Coach: Sounds like a plan. Good luck!!}                              \\ \hline
\textbf{Model Predictions} \\ \hline
\makecell[cl]{
\citet{gupta32021summarizing}: \texttt{\{Activity:Steps, Amount:{[}3 miles, \textcolor{red}{4000 steps}{]}, daytimes:\textcolor{red}{Tuesday}\}} \\
\citet{zhou-hc-22}: \texttt{\{Activity:Walk, Amount:3 miles, daytimes:\textcolor{red}{Friday}\}} \\
\textbf{Ours}: Walk 3 miles a day \textcolor{blue}{7 days a week}. \texttt{Instruction:Copy\{Days\}} \\
GPT-3.5-turbo: "Goal: 3 miles a day." - This goal is Specific (3 miles), Measurable (can be tracked by steps or distance), \\Achievable (based on the patient's previous performance), Relevant (to the patient's fitness objectives), and Time-bound \\(for the duration of the week). It's a well-defined and attainable goal. }\\
                                                             \hline
\end{tabular}
\end{adjustbox}
\caption{Qualitative examples of summarized goal from different models.}
\label{tab:qa1}
\end{table*}

As illustrated in Table~\ref{tab:qa2}, our Extended Point-wise $\mathcal{V}$-usable Information (PVI) for Generation measures unconventionality in the patient's response given the dialogue context. A significantly negative value indicates that the model struggles to anticipate the response, implying a higher degree of surprise in the patient's response. Conversely, a positive value suggests a more predictable and conventional response. Responses that either fail to directly address the coach's question (example \#1-3) or fall outside the domain of the conversation (example \#4, the patient intended to show her dog named King watching Youtube) are assigned a negative value. On the other hand, if the patient's response closely aligns with the coach's utterance, PVI will be positive. Interestingly, the response ``short of breath'' is unconventional while aligning with the coach's question, resulting in a moderately low value.

Table~\ref{tab:qa3} shows response generation outputs from different models. In Dialogue \#1, the model by~\cite{zhou-hc-22} failed to generate a coherent response due to the model's conditioning on the state of goal slot values while disregarding the critical contextual information (``set the same goals''). Interestingly, GPT-3.5 recognizes the patient's adherence to their goals but still prompts for a new goal definition. However, our model generates a coherent response that benefits from accurate Neuro-symbolic goal summarization. In Dialogue \#2, Our previous model failed to generate an appropriate response due to an error in updating the goal stage. The model incorrectly predicts the dialogue to be in the goal-setting stage, hence attempting to fill the empty attribute of the goal (e.g., days). In contrast, our NS generation does not utilize or predict stage information. Instead, it infers the dialogue progression, which is more likely to be in the goal implementation stage, based on a sequence of discrete units. It then focuses on the patient's concerns rather than completing the goals. Surprisingly, GPT3.5 failed to respond to the patient's concern, even when explicitly indicated in the prompt. The complete prompt for GPT3.5 dialogue generation can be found in the appendix.

\begin{table*}[!thb]
\begin{adjustbox}{width=\textwidth}
\begin{tabular}{ll}
\hline
\hline
\textbf{Dialogue Context$\,\to\,$Patient Response}                                                                                                         & \textbf{PVI}                        \\ \hline
\makecell[cl]{... Coach: and one last thing, how would you rate the goal? $\,\to\,$ Patient: yes and yes.}                            & \textcolor{red}{-0.11}            \\\hline

\makecell[cl]{... Coach: What would you like to set as your SMART goal this week? $\,\to\,$ Patient: Bam! Did you see the weekend?}                            & \textcolor{red}{-0.259}            \\\hline

\makecell[cl]{... Coach: Great and what days it will be? $\,\to\,$ Patient: Yes, but we can move it up to 8,000 steps.}                            & \textcolor{red}{-0.391}            \\\hline

\makecell[cl]{... Coach: It's also important to be kind to yourself. $\,\to\,$ Patient: This is King. Looking at Youtube.}                            & \textcolor{red}{-0.57}            \\\hline
\hline
\makecell[cl]{... Coach: Great would you be available at 9am or 11am? $\,\to\,$ Patient: 11am}                            & \textcolor{teal}{1.875}            \\\hline
\makecell[cl]{... Coach: How confident are you that you can reach your goal on a scale from 1 - 10? $\,\to\,$ Patient: 10. Absolutely.}                            & \textcolor{teal}{1.498}            \\\hline
\makecell[cl]{... Coach: what days would you like to do the elliptical? $\,\to\,$ Patient: Wednesday, Friday, and Saturday.}                            & \textcolor{teal}{1.096}            \\\hline
\makecell[cl]{... Coach: oh no! I'm sorry to hear that. What happened? $\,\to\,$ Patient: Short of breath.}                            & \textcolor{teal}{0.78}            \\\hline

\end{tabular}
\end{adjustbox}
\caption{Our Extended Point-wise $\mathcal{V}$-usable Information (PVI) for Generation measures unpredictability in the patient's response given the dialogue context. A large negative value shows the model's difficulty in predicting the response, indicating a higher degree of surprise in the patient's response. On the other hand, a positive value indicates a more predictable and conventional response.}
\label{tab:qa2}
\end{table*}

\begin{table*}[!thb]
\begin{adjustbox}{width=\textwidth}
\begin{tabular}{l}
       \\
\hline\hline
\textbf{Dialogue \#1} \\ \hline
\makecell[cl]{
Coach: Hi [Name], hope you're well. I forgot to message you yesterday to set your goal for Week 6, my apologies! \\Coach: You met your goal 5 out of 7 days last week, great job What would you like to do for this week?\\
Patient: Gd morning [Name], I am going to \textbf{set the same goals}. With a \textbf{7 confidence}.}                             \\ \hline
\textbf{Model Response} \\ \hline
\makecell[cl]{
\citet{zhou-hc-22}: That's great. And think about what activity would you like to set for your goal? Walks, stairs, etc\\
\textbf{Ours}: Ok, great! Can you think of any barriers that might prevent your goal? \\
GPT-3.5-turbo: That's fantastic to hear that you're sticking to your goals! Let's make sure it's a SMART goal for Week 6.\\ Can you provide more details about the goal, like the specific number of steps or any changes in frequency, time, or location?}\\
                                                             \hline

                                                             \textbf{Dialogue \#2} \\ \hline
\makecell[cl]{
Coach: Hi [Name]! Hope you had a good weekend and were able to stay cool/dry! How are you feeling about your 5,000 steps goal this week?\\
Patient: Good morning. I'm feeling good about it. I will start today (Monday) this weekend I had a party on Saturday and rested on Sunday.\\
Coach: Sounds like a fun weekend! I'm excited to see your steps for today and the rest of the week!\\
Patient: Loved "Sounds like a fun weekend! I\'m excited to see your steps for today and the rest of the week!"\\
Coach: You are totally crushing your goal this week! Great job! Keep it up! \\
Patient: Thank you.\\
Coach: Reminder to sync your Fitbit!\\
Patient: What do you mean? Are you unable to track me?}                             \\ \hline
\textbf{Model Response} \\ \hline
\makecell[cl]{
\citet{zhou-hc-22}: What days? \\
\textbf{Ours}: Oh, Just wanted to remind you. \\
GPT-3.5-turbo: That's great to hear you're feeling good about your goal! How many days this week do you plan to aim for those 5,000 steps?}\\
                                                             \hline

\end{tabular}
\end{adjustbox}
\caption{Qualitative examples and error analysis for response generation.}
\label{tab:qa3}
\end{table*}

\section{Conclusion}

We propose a neuro-symbolic goal summarization approach that (i) summarizes the goal as effectively as possible, given the current week's dialogue, and (ii) generates an executable instruction on modifying the summarized goal by referencing the previous goal. We also build a text-units-text dialogue generation approach that considers as input a sequence of discrete units symbolizing the dialogue history. Our models outperform previous state-of-the-art while eliminating the need for a pre-designed schema or corresponding annotations. This significantly minimizes the annotation labor cost and development cycle, benefiting healthcare and other resource-constrained scenarios. The framework also exhibits the potential for generalization to other applications, including developing dialogue systems in patient education upon discharge or consultations regarding behavioral change issues. However, there remains to be a disparity between the responses generated by the model and those of expert human levels. Given the nature of data limits and sensitivity in healthcare, generating human-level, fair, and faithful outputs while ensuring the model's interpretability is a significant and challenging direction for future exploration.
%\vspace{5cm}
%\section{Acknowledgements}

%This work is supported by the National Science Foundation under Grant IIS-1838770.
\section{Acknowledgments}
The work is supported by the National Science Foundation under Grant IIS-1838770.

%\nocite{*}

%\newpage
\section{Bibliographical References}\label{sec:reference}

\bibliographystyle{lrec-coling2024-natbib}
\bibliography{lrec-coling2024-example}

\begin{thebibliography}{0}
\expandafter\ifx\csname natexlab\endcsname\relax\def\natexlab#1{#1}\fi

\end{thebibliography}


\begin{thebibliography}{46}
\expandafter\ifx\csname natexlab\endcsname\relax\def\natexlab#1{#1}\fi

\bibitem[{Althoff et~al.(2016)Althoff, Clark, and
  Leskovec}]{consolalthoff-etal-2016-large}
Tim Althoff, Kevin Clark, and Jure Leskovec. 2016.
\newblock \href {https://doi.org/10.1162/tacl_a_00111} {Large-scale analysis of
  counseling conversations: An application of natural language processing to
  mental health}.
\newblock \emph{Transactions of the Association for Computational Linguistics},
  4:463--476.

\bibitem[{Arabshahi et~al.(2021)Arabshahi, Lee, Gawarecki, Mazaitis, Azaria,
  and Mitchell}]{nsconv}
Forough Arabshahi, Jennifer Lee, Mikayla Gawarecki, Kathryn Mazaitis, Amos
  Azaria, and Tom Mitchell. 2021.
\newblock Conversational neuro-symbolic commonsense reasoning.
\newblock In \emph{Proceedings of the AAAI Conference on Artificial
  Intelligence}, volume~35, pages 4902--4911.

\bibitem[{Budzianowski et~al.(2018)Budzianowski, Wen, Tseng, Casanueva, Ultes,
  Ramadan, and Ga{\v{s}}i{\'c}}]{tod2budzianowski-etal-2018-multiwoz}
Pawe{\l} Budzianowski, Tsung-Hsien Wen, Bo-Hsiang Tseng, I{\~n}igo Casanueva,
  Stefan Ultes, Osman Ramadan, and Milica Ga{\v{s}}i{\'c}. 2018.
\newblock \href {https://doi.org/10.18653/v1/D18-1547} {{M}ulti{WOZ} - a
  large-scale multi-domain {W}izard-of-{O}z dataset for task-oriented dialogue
  modelling}.
\newblock In \emph{Proceedings of the 2018 Conference on Empirical Methods in
  Natural Language Processing}, pages 5016--5026, Brussels, Belgium.
  Association for Computational Linguistics.

\bibitem[{Butterworth et~al.(2006)Butterworth, Linden, McClay, and
  Leo}]{hc4butterworth2006effect}
Susan Butterworth, Ariel Linden, Wende McClay, and Michael~C Leo. 2006.
\newblock Effect of motivational interviewing-based health coaching on
  employees' physical and mental health status.
\newblock \emph{Journal of occupational health psychology}, 11(4):358.

\bibitem[{Chaix et~al.(2020)Chaix, Guillemass{\'e}, Nectoux, Delamon, Brouard
  et~al.}]{sms3chaix2020vik}
Benjamin Chaix, Arthur Guillemass{\'e}, Pierre Nectoux, Guillaume Delamon,
  Beno{\^\i}t Brouard, et~al. 2020.
\newblock Vik: a chatbot to support patients with chronic diseases.
\newblock \emph{Health}, 12(07):804.

\bibitem[{De~Raedt et~al.(2019)De~Raedt, Manhaeve, Dumancic, Demeester, and
  Kimmig}]{deraedt2019neurosymbolic}
Luc De~Raedt, Robin Manhaeve, Sebastijan Dumancic, Thomas Demeester, and
  Angelika Kimmig. 2019.
\newblock Neurosymbolic= neural+ logical+ probabilistic.
\newblock In \emph{NeSy'19@ IJCAI, the 14th International Workshop on
  NeuralSymbolic Learning and Reasoning}.

\bibitem[{\disambiguate{Tianyi Zhang}{Zhang}{Tianyi Zhang}
  et~al.(2020)\disambiguate{Tianyi Zhang}{Zhang}{Tianyi Zhang}, Kishore, Wu,
  Weinberger, and Artzi}]{bert-score}
\disambiguate{Tianyi Zhang}{Zhang}{Tianyi Zhang}, Varsha Kishore, Felix Wu,
  Kilian~Q. Weinberger, and Yoav Artzi. 2020.
\newblock \href {https://openreview.net/forum?id=SkeHuCVFDr} {Bertscore:
  Evaluating text generation with bert}.
\newblock In \emph{International Conference on Learning Representations}.

\bibitem[{Dong and Lapata(2018)}]{dong-lapata-2018-coarse}
Li~Dong and Mirella Lapata. 2018.
\newblock \href {https://doi.org/10.18653/v1/P18-1068} {Coarse-to-fine decoding
  for neural semantic parsing}.
\newblock In \emph{Proceedings of the 56th Annual Meeting of the Association
  for Computational Linguistics (Volume 1: Long Papers)}, pages 731--742,
  Melbourne, Australia. Association for Computational Linguistics.

\bibitem[{Doran(1981)}]{doran1981there}
George~T Doran. 1981.
\newblock There's a {SMART} way to write management’s goals and objectives.
\newblock \emph{Management review}, 70(11):35--36.

\bibitem[{Ethayarajh et~al.(2022)Ethayarajh, Choi, and Swayamdipta}]{PVI}
Kawin Ethayarajh, Yejin Choi, and Swabha Swayamdipta. 2022.
\newblock Understanding dataset difficulty with $\mathcal{V}$-usable
  information.
\newblock In \emph{Proceedings of the 39th International Conference on Machine
  Learning}, volume 162 of \emph{Proceedings of Machine Learning Research},
  pages 5988--6008. PMLR.

\bibitem[{Feng et~al.(2020)Feng, Zheng, Liu, Greenspan, and
  Zhu}]{feng-etal-2020-exploring}
Yufei Feng, Zi{'}ou Zheng, Quan Liu, Michael Greenspan, and Xiaodan Zhu. 2020.
\newblock \href {https://doi.org/10.18653/v1/2020.coling-main.101} {Exploring
  end-to-end differentiable natural logic modeling}.
\newblock In \emph{Proceedings of the 28th International Conference on
  Computational Linguistics}, pages 1172--1185, Barcelona, Spain (Online).
  International Committee on Computational Linguistics.

\bibitem[{Fitzpatrick et~al.(2017)Fitzpatrick, Darcy, and
  Vierhile}]{sms2fitzpatrick2017delivering}
Kathleen~Kara Fitzpatrick, Alison Darcy, and Molly Vierhile. 2017.
\newblock Delivering cognitive behavior therapy to young adults with symptoms
  of depression and anxiety using a fully automated conversational agent
  (woebot): a randomized controlled trial.
\newblock \emph{JMIR mental health}, 4(2):e7785.

\bibitem[{Garcez et~al.(2018)Garcez, Gori, Lamb, and
  Serafini}]{garcez2018neurosymbolic}
Artur~d'Avila Garcez, Marco Gori, Luise Lamb, and Luciano Serafini. 2018.
\newblock Neural-symbolic learning and reasoning: A survey and taxonomy.
\newblock \emph{IEEE Transactions on Neural Networks and Learning Systems},
  29(10):4869--4881.

\bibitem[{Ghorob(2013)}]{hc1ghorob2013supplement}
Amireh Ghorob. 2013.
\newblock Supplement: Health coaching: Teaching patients to fish.
\newblock \emph{Family practice management}, 20(3):40--42.

\bibitem[{Gupta et~al.(2020{\natexlab{a}})Gupta, Di~Eugenio, Ziebart, Baiju,
  Liu, Gerber, Sharp, Nabulsi, and Smart}]{gupta1-etal-2020-human}
Itika Gupta, Barbara Di~Eugenio, Brian Ziebart, Aiswarya Baiju, Bing Liu, Ben
  Gerber, Lisa Sharp, Nadia Nabulsi, and Mary Smart. 2020{\natexlab{a}}.
\newblock \href {https://aclanthology.org/2020.sigdial-1.30} {Human-human
  health coaching via text messages: Corpus, annotation, and analysis}.
\newblock In \emph{Proceedings of the 21th Annual Meeting of the Special
  Interest Group on Discourse and Dialogue}, pages 246--256, 1st virtual
  meeting. Association for Computational Linguistics.

\bibitem[{Gupta et~al.(2020{\natexlab{b}})Gupta, Di~Eugenio, Ziebart, Liu,
  Gerber, and Sharp}]{gupta22020goal}
Itika Gupta, Barbara Di~Eugenio, Brian Ziebart, Bing Liu, Ben Gerber, and Lisa
  Sharp. 2020{\natexlab{b}}.
\newblock Goal summarization for human-human health coaching dialogues.
\newblock In \emph{The Thirty-Third International Flairs Conference}.

\bibitem[{Gupta et~al.(2021)Gupta, Di~Eugenio, Ziebart, Liu, Gerber, and
  Sharp}]{gupta32021summarizing}
Itika Gupta, Barbara Di~Eugenio, Brian~D Ziebart, Bing Liu, Ben~S Gerber, and
  Lisa~K Sharp. 2021.
\newblock Summarizing behavioral change goals from sms exchanges to support
  health coaches.
\newblock In \emph{Proceedings of the 22nd Annual Meeting of the Special
  Interest Group on Discourse and Dialogue}, pages 276--289.

\bibitem[{Hosseini-Asl et~al.(2020)Hosseini-Asl, McCann, Wu, Yavuz, and
  Socher}]{tode2e3hosseini2020simple}
Ehsan Hosseini-Asl, Bryan McCann, Chien-Sheng Wu, Semih Yavuz, and Richard
  Socher. 2020.
\newblock A simple language model for task-oriented dialogue.
\newblock \emph{Advances in Neural Information Processing Systems},
  33:20179--20191.

\bibitem[{Jokinen and McTear(2009)}]{jokinen09}
Kristiina Jokinen and Michael McTear. 2009.
\newblock Spoken dialogue systems.
\newblock \emph{Synthesis Lectures on Human Language Technologies},
  2(1):1--151.

\bibitem[{Kangovi et~al.(2014)Kangovi, Barg, Carter, Levy, Sellman, Long, and
  Grande}]{ses1kangovi2014challenges}
Shreya Kangovi, Frances~K Barg, Tamala Carter, Kathryn Levy, Jeffrey Sellman,
  Judith~A Long, and David Grande. 2014.
\newblock Challenges faced by patients with low socioeconomic status during the
  post-hospital transition.
\newblock \emph{Journal of general internal medicine}, 29(2):283--289.

\bibitem[{Kivel{\"a} et~al.(2014)Kivel{\"a}, Elo, Kyng{\"a}s, and
  K{\"a}{\"a}ri{\"a}inen}]{hc3kivela2014effects}
Kirsi Kivel{\"a}, Satu Elo, Helvi Kyng{\"a}s, and Maria K{\"a}{\"a}ri{\"a}inen.
  2014.
\newblock The effects of health coaching on adult patients with chronic
  diseases: a systematic review.
\newblock \emph{Patient education and counseling}, 97(2):147--157.

\bibitem[{Kocielnik et~al.(2018)Kocielnik, Xiao, Avrahami, and
  Hsieh}]{sms5kocielnik2018reflection}
Rafal Kocielnik, Lillian Xiao, Daniel Avrahami, and Gary Hsieh. 2018.
\newblock Reflection companion: a conversational system for engaging users in
  reflection on physical activity.
\newblock \emph{Proceedings of the ACM on Interactive, Mobile, Wearable and
  Ubiquitous Technologies}, 2(2):1--26.

\bibitem[{Lakhotia et~al.(2021)Lakhotia, Kharitonov, Hsu, Adi, Polyak, Bolte,
  Nguyen, Copet, Baevski, Mohamed, and Dupoux}]{unit2}
Kushal Lakhotia, Eugene Kharitonov, Wei-Ning Hsu, Yossi Adi, Adam Polyak,
  Benjamin Bolte, Tu-Anh Nguyen, Jade Copet, Alexei Baevski, Abdelrahman
  Mohamed, and Emmanuel Dupoux. 2021.
\newblock \href {https://doi.org/10.1162/tacl_a_00430} {On generative spoken
  language modeling from raw audio}.
\newblock \emph{Transactions of the Association for Computational Linguistics},
  9:1336--1354.

\bibitem[{Lamb et~al.(2019)Lamb, Goyal, Zhang, Zhang, Courville, and
  Bengio}]{Lamb2019}
Alex Lamb, Anirudh Goyal, Yunyang Zhang, Saizheng Zhang, Aaron Courville, and
  Yoshua Bengio. 2019.
\newblock Neuro-symbolic transformers for text classification: A case for
  symbolic reasoning in deep learning.
\newblock In \emph{NeurIPS}.

\bibitem[{Lee et~al.(2022)Lee, Chen, Wang, Gu, Popuri, Ma, Polyak, Adi, He,
  Tang, Pino, and Hsu}]{unit1}
Ann Lee, Peng-Jen Chen, Changhan Wang, Jiatao Gu, Sravya Popuri, Xutai Ma, Adam
  Polyak, Yossi Adi, Qing He, Yun Tang, Juan Pino, and Wei-Ning Hsu. 2022.
\newblock \href {https://doi.org/10.18653/v1/2022.acl-long.235} {Direct
  speech-to-speech translation with discrete units}.
\newblock In \emph{Proceedings of the 60th Annual Meeting of the Association
  for Computational Linguistics (Volume 1: Long Papers)}, pages 3327--3339,
  Dublin, Ireland. Association for Computational Linguistics.

\bibitem[{Lin(2004)}]{rouge}
Chin-Yew Lin. 2004.
\newblock \href {https://aclanthology.org/W04-1013} {{ROUGE}: A package for
  automatic evaluation of summaries}.
\newblock In \emph{Text Summarization Branches Out}, pages 74--81, Barcelona,
  Spain. Association for Computational Linguistics.

\bibitem[{Ma et~al.(2020)Ma, Sap, Rashkin, and
  Choi}]{ppl-ma-etal-2020-powertransformer}
Xinyao Ma, Maarten Sap, Hannah Rashkin, and Yejin Choi. 2020.
\newblock \href {https://doi.org/10.18653/v1/2020.emnlp-main.602}
  {{P}ower{T}ransformer: Unsupervised controllable revision for biased language
  correction}.
\newblock In \emph{Proceedings of the 2020 Conference on Empirical Methods in
  Natural Language Processing (EMNLP)}, pages 7426--7441, Online. Association
  for Computational Linguistics.

\bibitem[{Mao et~al.(2019)Mao, Gan, Kohli, Tenenbaum, and Wu}]{Mao2019}
Jiayuan Mao, Chuang Gan, Pushmeet Kohli, Joshua~B Tenenbaum, and Jiajun Wu.
  2019.
\newblock The neuro-symbolic concept learner: Interpreting scenes, words, and
  sentences from natural supervision.
\newblock \emph{ICLR}.

\bibitem[{Mohan et~al.(2020)Mohan, Venkatakrishnan, and
  Hartzler}]{sms4mohan2020designing}
Shiwali Mohan, Anusha Venkatakrishnan, and Andrea~L Hartzler. 2020.
\newblock Designing an ai health coach and studying its utility in promoting
  regular aerobic exercise.
\newblock \emph{ACM Transactions on Interactive Intelligent Systems (TiiS)},
  10(2):1--30.

\bibitem[{Mrk{\v{s}}i{\'c} et~al.(2017)Mrk{\v{s}}i{\'c}, {\'O}~S{\'e}aghdha,
  Wen, Thomson, and Young}]{tod3mrkvsic2016neural}
Nikola Mrk{\v{s}}i{\'c}, Diarmuid {\'O}~S{\'e}aghdha, Tsung-Hsien Wen, Blaise
  Thomson, and Steve Young. 2017.
\newblock \href {https://doi.org/10.18653/v1/P17-1163} {Neural belief tracker:
  Data-driven dialogue state tracking}.
\newblock In \emph{Proceedings of the 55th Annual Meeting of the Association
  for Computational Linguistics (Volume 1: Long Papers)}, pages 1777--1788,
  Vancouver, Canada. Association for Computational Linguistics.

\bibitem[{Papineni et~al.(2002)Papineni, Roukos, Ward, and
  Zhu}]{bleu-papineni-etal-2002-bleu}
Kishore Papineni, Salim Roukos, Todd Ward, and Wei-Jing Zhu. 2002.
\newblock \href {https://doi.org/10.3115/1073083.1073135} {{B}leu: a method for
  automatic evaluation of machine translation}.
\newblock In \emph{Proceedings of the 40th Annual Meeting of the Association
  for Computational Linguistics}, pages 311--318, Philadelphia, Pennsylvania,
  USA. Association for Computational Linguistics.

\bibitem[{Peng et~al.(2021)Peng, Li, Li, Shayandeh, Liden, and
  Gao}]{tode2e4peng2020soloist}
Baolin Peng, Chunyuan Li, Jinchao Li, Shahin Shayandeh, Lars Liden, and
  Jianfeng Gao. 2021.
\newblock \href {https://doi.org/10.1162/tacl_a_00399} {Soloist: Building task
  bots at scale with transfer learning and machine teaching}.
\newblock \emph{Transactions of the Association for Computational Linguistics},
  9:807--824.

\bibitem[{Pleiss et~al.(2020)Pleiss, Zhang, Elenberg, and Weinberger}]{AUM}
Geoff Pleiss, Tianyi Zhang, Ethan Elenberg, and Kilian~Q Weinberger. 2020.
\newblock \href
  {https://proceedings.neurips.cc/paper/2020/file/c6102b3727b2a7d8b1bb6981147081ef-Paper.pdf}
  {Identifying mislabeled data using the area under the margin ranking}.
\newblock In \emph{Advances in Neural Information Processing Systems},
  volume~33, pages 17044--17056. Curran Associates, Inc.

\bibitem[{Radford et~al.(2019)Radford, Wu, Child, Luan, Amodei, Sutskever
  et~al.}]{gpt2-radford2019language}
Alec Radford, Jeffrey Wu, Rewon Child, David Luan, Dario Amodei, Ilya
  Sutskever, et~al. 2019.
\newblock Language models are unsupervised multitask learners.
\newblock \emph{OpenAI blog}, 1(8):9.

\bibitem[{Raffel et~al.(2020)Raffel, Shazeer, Roberts, Lee, Narang, Matena,
  Zhou, Li, and Liu}]{t5-raffel2019exploring}
Colin Raffel, Noam Shazeer, Adam Roberts, Katherine Lee, Sharan Narang, Michael
  Matena, Yanqi Zhou, Wei Li, and Peter~J. Liu. 2020.
\newblock \href {http://jmlr.org/papers/v21/20-074.html} {Exploring the limits
  of transfer learning with a unified text-to-text transformer}.
\newblock \emph{Journal of Machine Learning Research}, 21(140):1--67.

\bibitem[{Schulman et~al.(2017)Schulman, Wolski, Dhariwal, Radford, and
  Klimov}]{PPO}
John Schulman, Filip Wolski, Prafulla Dhariwal, Alec Radford, and Oleg Klimov.
  2017.
\newblock \href {http://arxiv.org/abs/1707.06347} {Proximal policy optimization
  algorithms}.

\bibitem[{Sharma et~al.(2021)Sharma, Lin, Miner, Atkins, and
  Althoff}]{emp-rewriting10.1145/3442381.3450097}
Ashish Sharma, Inna~W. Lin, Adam~S. Miner, David~C. Atkins, and Tim Althoff.
  2021.
\newblock \href {https://doi.org/10.1145/3442381.3450097} {\emph{Towards
  Facilitating Empathic Conversations in Online Mental Health Support: A
  Reinforcement Learning Approach}}, page 194–205. Association for Computing
  Machinery, New York, NY, USA.

\bibitem[{Shen et~al.(2020)Shen, Welch, Mihalcea, and
  P{\'e}rez-Rosas}]{consol2shen-etal-2020-counseling}
Siqi Shen, Charles Welch, Rada Mihalcea, and Ver{\'o}nica P{\'e}rez-Rosas.
  2020.
\newblock \href {https://aclanthology.org/2020.sigdial-1.2} {Counseling-style
  reflection generation using generative pretrained transformers with augmented
  context}.
\newblock In \emph{Proceedings of the 21th Annual Meeting of the Special
  Interest Group on Discourse and Dialogue}, pages 10--20, 1st virtual meeting.
  Association for Computational Linguistics.

\bibitem[{Swayamdipta et~al.(2020)Swayamdipta, Schwartz, Lourie, Wang,
  Hajishirzi, Smith, and Choi}]{carto}
Swabha Swayamdipta, Roy Schwartz, Nicholas Lourie, Yizhong Wang, Hannaneh
  Hajishirzi, Noah~A. Smith, and Yejin Choi. 2020.
\newblock \href {https://doi.org/10.18653/v1/2020.emnlp-main.746} {Dataset
  cartography: Mapping and diagnosing datasets with training dynamics}.
\newblock In \emph{Proceedings of the 2020 Conference on Empirical Methods in
  Natural Language Processing (EMNLP)}, pages 9275--9293, Online. Association
  for Computational Linguistics.

\bibitem[{Thackeray et~al.(2004)Thackeray, Merrill, and
  Neiger}]{ses2thackeray2004disparities}
Rosemary Thackeray, Ray~M Merrill, and Brad~L Neiger. 2004.
\newblock Disparities in diabetes management practice between racial and ethnic
  groups in the united states.
\newblock \emph{The Diabetes Educator}, 30(4):665--675.

\bibitem[{Thom et~al.(2016)Thom, Wolf, Gardner, DeVore, Lin, Ma, Ibarra-Castro,
  and Saba}]{hc2thom2016qualitative}
David~H Thom, Jessica Wolf, Heather Gardner, Denise DeVore, Michael Lin, Andy
  Ma, Ana Ibarra-Castro, and George Saba. 2016.
\newblock A qualitative study of how health coaches support patients in making
  health-related decisions and behavioral changes.
\newblock \emph{The Annals of Family Medicine}, 14(6):509--516.

\bibitem[{Wen et~al.(2015)Wen, Ga{\v{s}}i{\'c}, Mrk{\v{s}}i{\'c}, Su, Vandyke,
  and Young}]{tod4wen-etal-2015-semantically}
Tsung-Hsien Wen, Milica Ga{\v{s}}i{\'c}, Nikola Mrk{\v{s}}i{\'c}, Pei-Hao Su,
  David Vandyke, and Steve Young. 2015.
\newblock \href {https://doi.org/10.18653/v1/D15-1199} {Semantically
  conditioned {LSTM}-based natural language generation for spoken dialogue
  systems}.
\newblock In \emph{Proceedings of the 2015 Conference on Empirical Methods in
  Natural Language Processing}, pages 1711--1721, Lisbon, Portugal. Association
  for Computational Linguistics.

\bibitem[{Williams et~al.(2016)Williams, Raux, and
  Henderson}]{tod1williams2016dialog}
Jason~D Williams, Antoine Raux, and Matthew Henderson. 2016.
\newblock The dialog state tracking challenge series: A review.
\newblock \emph{Dialogue \& Discourse}, 7(3):4--33.

\bibitem[{Wolf et~al.(2020)Wolf, Debut, Sanh, Chaumond, Delangue, Moi, Cistac,
  Rault, Louf, Funtowicz, Davison, Shleifer, von Platen, Ma, Jernite, Plu, Xu,
  Le~Scao, Gugger, Drame, Lhoest, and
  Rush}]{hugginghttps://doi.org/10.48550/arxiv.1910.03771}
Thomas Wolf, Lysandre Debut, Victor Sanh, Julien Chaumond, Clement Delangue,
  Anthony Moi, Pierric Cistac, Tim Rault, Remi Louf, Morgan Funtowicz, Joe
  Davison, Sam Shleifer, Patrick von Platen, Clara Ma, Yacine Jernite, Julien
  Plu, Canwen Xu, Teven Le~Scao, Sylvain Gugger, Mariama Drame, Quentin Lhoest,
  and Alexander Rush. 2020.
\newblock \href {https://doi.org/10.18653/v1/2020.emnlp-demos.6} {Transformers:
  State-of-the-art natural language processing}.
\newblock In \emph{Proceedings of the 2020 Conference on Empirical Methods in
  Natural Language Processing: System Demonstrations}, pages 38--45, Online.
  Association for Computational Linguistics.

\bibitem[{Xu et~al.(2020)Xu, Zhao, Song, Stewart, and Ermon}]{pvi-thoery}
Yilun Xu, Shengjia Zhao, Jiaming Song, Russell Stewart, and Stefano Ermon.
  2020.
\newblock \href {https://openreview.net/forum?id=r1eBeyHFDH} {A theory of
  usable information under computational constraints}.
\newblock In \emph{International Conference on Learning Representations}.

\bibitem[{Zhou et~al.(2022)Zhou, Di~Eugenio, Ziebart, Sharp, Liu, Gerber,
  Agadakos, and Yadav}]{zhou-hc-22}
Yue Zhou, Barbara Di~Eugenio, Brian Ziebart, Lisa Sharp, Bing Liu, Ben Gerber,
  Nikolaos Agadakos, and Shweta Yadav. 2022.
\newblock \href {https://aclanthology.org/2022.coling-1.58} {Towards enhancing
  health coaching dialogue in low-resource settings}.
\newblock In \emph{Proceedings of the 29th International Conference on
  Computational Linguistics}, pages 694--706, Gyeongju, Republic of Korea.
  International Committee on Computational Linguistics.

\end{thebibliography}

%\section{Language Resource References}
\label{lr:ref}
\bibliographystylelanguageresource{lrec-coling2024-natbib}
\bibliographylanguageresource{languageresource}

\newpage
%\section*{Appendix}
\appendix

\section{Instruction Set}

The full set of instruction on goal editing based on the reference is shown in Table~\ref{tab:set}.

\begin{table}[ht]
\centering
\begin{adjustbox}{width=\columnwidth}

% \usepackage[normalem]{ulem}

\begin{tabular}{ll}
\hline\hline
Instruction  & Explanation\\ \hline
\texttt{" "} & Pass \\
\texttt{Copy \{Times\}} & Copy times, e.g., ``after 8 pm'' \\
\texttt{Copy \{Days\}} & Copy days, e.g., ``Mon-Fri'' \\
\texttt{Copy \{Num\}} & Copy number, e.g., ``3000 steps'' \\
\texttt{Add \{Num\}} & Add number from previous goal.  \\
\texttt{Add \{Days\}} & Add days from previous goal.  \\
\texttt{Copy \{All\}} & Copy all info from previous.  \\
\hline
\end{tabular}

\end{adjustbox}
\caption{The executable instructions we used in the neuro-symbolic goal summarization.}
\label{tab:set}
\end{table}

\section{Prompts}
We show the templates we use when prompting the language model for goal summarization and response generation.
\paragraph{Prompt for Goal Summarization:} \texttt{Summarize the SMART goal discussed in the health coaching dialogue. The goal attributes include details such as the activity (e.g., walking), the quantity of the activity (e.g., 4000 steps), the schedule (e.g., Monday-Friday, after 4 pm), locations, and more:/n}

\paragraph{Prompt for Response Generation:} \texttt{As a health coach, your task is to refine the patient's goal into a SMART goal. Ask for details like frequency, time, duration, location, and confidence level, but only one aspect at a time. After establishing the goal, monitor the patient's progress and keep them engaged. Always address their concerns concisely. Now, consider the following dialogue context and formulate your response:/n}

\section{Quantitative Results of LLMs}
Due to resource constraints, we run the goal summarization and dialogue generation on 100 randomly selected examples with GPT-3.5-turbo. We show the results in Table~\ref{q35}.

\begin{table}[ht]
\centering
\begin{adjustbox}{width=0.65\columnwidth}

% \usepackage[normalem]{ulem}

\begin{tabular}{ll}
\hline\hline
\textbf{Response Generation} &       \\\hline
PPL                 & 23.6  \\
BLEU                & 14.21 \\
BertScore           & 85.62 \\ \hline
\textbf{Goal Summarization}  &       \\\hline
Semantic Frame Acc  & 41.0 \\ \hline   
\end{tabular}

\end{adjustbox}
\caption{Performance of GPT-3.5-turbo on sampled response generation and goal summarization data.}
\label{q35}
\end{table}

\section{Training Details}
All the following models use Huggingface Transformers Library ~\citep{hugginghttps://doi.org/10.48550/arxiv.1910.03771}. The hyperparameters are not extensively fine-tuned.

 For the goal summarizer, we use T5-base as the model backbone. To mitigate inefficient sampling, we manually annotate 40 positive examples, i.e.,  the sequences of instructions and partial goals that result in ground truth. The model was first fine-tuned with dialogue-to-goal pairs and then with the 40 examples we labeled. Finally, we contrast the sampled negative examples (the ones that fail) with the positive examples to update the gradient. 
 
 We also use two T5-base models to train our PVI-generation metrics. Model $g$ was trained with the context mapping to the patient response for two epochs, while $g'$ was trained with an empty context with one epoch. 
 
For dialogue generation, we choose $k$ = 15 for deriving the discrete units. We use GPT-2-large as the model backbone with a max sequence length set to 128 since the dialogue history has been symbolized as discrete units. The model was trained for 7.0 epochs with a learning rate of 1e-4, with a batch size of 16. We use sampling during decoding with top-k set to 40 and top-p set to 1.

\end{document}